\documentclass[runningheads]{llncs}

\usepackage{eccv}

\usepackage{eccvabbrv}

\usepackage{graphicx}
\usepackage{booktabs}
\usepackage{cite}
\usepackage{amsmath,amssymb,amsfonts}
\usepackage{algorithmic}
\usepackage{bbold}  
\usepackage{comment}
\usepackage{textcomp}
\usepackage{siunitx}
\usepackage{tabularx}
\usepackage{makecell}
\usepackage{multirow}
\usepackage{cuted}
\usepackage{capt-of}
\usepackage{rotating}
\usepackage{bigstrut} 
\usepackage{pifont}
\usepackage{wasysym}
\usepackage{environ}

\usepackage[accsupp]{axessibility}  

\usepackage{hyperref}

\usepackage{orcidlink}

\NewEnviron{gather1}[1][1]{%
  \begin{equation}\label{equ:confidence_level}
  \scalebox{#1}{$\begin{gathered}\BODY\end{gathered}$}
  \end{equation}
}

\begin{document}

\title{Reliable Probabilistic Human Trajectory Prediction for Autonomous Applications} 

\titlerunning{Reliable Probabilistic Human Trajectory Prediction}

\author{Manuel Hetzel\inst{1}\orcidlink{0000-0002-6737-8935}\thanks{The authors contributed equally.} \and
Hannes Reichert\inst{1}\orcidlink{0000-0002-8710-9409}\textsuperscript{*} \and
Konrad Doll\inst{1}\orcidlink{0000-0002-3746-2319}\and
Bernhard Sick\inst{2}\orcidlink{0000-0001-9467-656X}}

\authorrunning{M.~Hetzel et al.}

\institute{University of Applied Sciences Aschaffenburg, Germany \\ \email{\{manuel.hetzel, hannes.reichert, konrad.doll\}@th-ab.de} \and
University of Kassel, Germany \\ \email{\{bsick\}@uni-kassel.de}
}

\maketitle

\begin{abstract} 

Autonomous systems, like vehicles or robots, require reliable, accurate, fast, resource-efficient, scalable, and low-latency trajectory predictions to get initial knowledge about future locations and movements of surrounding objects for safe human-machine interaction. Furthermore, they need to know the uncertainty of the predictions for risk assessment to provide safe path planning. This paper presents a lightweight method to address these requirements, combining Long Short-Term Memory and Mixture Density Networks. Our method predicts probability distributions, including confidence level estimations for positional uncertainty to support subsequent risk management applications and runs on a low-power embedded platform. We discuss essential requirements for human trajectory prediction in autonomous vehicle applications and demonstrate our method's performance using multiple traffic-related datasets. Furthermore, we explain reliability and sharpness metrics and show how important they are to guarantee the correctness and robustness of a model's predictions and uncertainty assessments. These essential evaluations have so far received little attention for no good reason. Our approach focuses entirely on real-world applicability. Verifying prediction uncertainties and a model's reliability are central to autonomous real-world applications. Our framework and code are available at: \url{https://github.com/kav-institute/mdn_trajectory_forecasting}.


\end{abstract}
\section{\large Introduction}\label{sec_introduction}

\subsection{Motivation}\label{sec_motivation}
Predicting the future residence areas of humans with uncertainties is necessary to enable Autonomous Machines (AM) like Autonomous Vehicles (AV), forklifts, or freight transporters. This article focuses on Human Trajectory Prediction (HTP) for autonomous vehicles. AVs will share the road with Vulnerable Road Users (VRU). To guarantee safe interaction, AVs not only have to be able to perceive their current environment, but they must also anticipate the future movements of all surrounding participants. While AVs can share their planned paths, humans' future trajectories must be estimated. Human movements are in-deterministic by nature, and movement paths could be changed quickly, resulting in error-prone predictions. As humans tend to overestimate their abilities, a cognitive bias known as the Dunning-Kruger effect~\cite{kruger}, neural networks also tend to be under- or overconfident - they predict with low/high confidence even when they are right/wrong~\cite{calibration}.
Therefore, reviewing neural network predictions regarding reliability and robustness is highly important. Current research on HTP primarily aims to use deterministic evaluation. $N$ discrete hypotheses are generated and evaluated by testing these against ground-truth data for accuracy. Several valid application areas exist for said method, like human interaction and motion flow analysis or motion target determination in public places. However, these procedures are unsuitable for autonomous applications; for deterministic evaluation at execution time, no ground truth data is available, and for probabilistic hypotheses, the commonly used evaluation metrics Average Displacement Error (ADE), Final displacement Error (FDE), and Negative Log-likelihood (NLL) can not evaluate the reliability and robustness. If a prediction states that a pedestrian has a 90\% probability of being within a specified area, the prediction should occur 90\% of the time it is made. Otherwise, downstream tasks like path-planing cannot trust these predictions. Therefore, as the minimum requirement, discrete hypotheses with reliably calibrated probabilities or better confidence areas where humans will reside with a certain likelihood are mandatory to address uncertainty in HTP tasks. 

Moreover, in practical use, HTP algorithms must share the available processing resources and scale in multiple flexible aspects. Our approach aims to cover all aspects. This creates some limitations in terms of architectural choice and distinguishes it from most other methods in HTP research. For this reason, generally known requirements are addressed and discussed below to clarify the background of our work.

\subsection{Requirements for HTP in Autonomous Machine Applications}\label{sec_requirements}
We summarize essential requirements for trajectory prediction methods in autonomous machine applications and derive design decisions. The primary goal is to enable safe interaction between humans and AMs.

\textbf{Uncertainty Modeling:} It reflects the inherent unpredictability and variability in real-world systems and processes that forecasts attempt to predict. Uncertainty can arise from various sources (incomplete or noisy data, stochastic events, and complexity of the systems being modeled). Modeling uncertainties in HTP is significant for risk assessment. Therefore, a model should provide additional information to address the trajectory prediction uncertainty. This can be achieved by supplying confidence areas or scores during inference.

\textbf{Reliability and Confidence:} The reliability of uncertainty estimates in predictions provides decision-making processes with a measure of confidence. Reliability in this context refers to the trustworthiness of the uncertainty information associated with a prediction. \cite{zernetsch_holistic} introduced a stochastic calibration metric to evaluate the reliability of probabilistic models. This work recaptures important results and demonstrates its importance for autonomous applications. If a model produces under- or overconfident predictions, it poses a significant risk. Unreliable predictions are not credible and cannot be used for critical path-planning tasks, underlining the need for improved HTP evaluation. Therefore, in Chapter~\ref{sec_method}, we introduced two new metrics to measure \textit{Reliability} and \textit{Sharpness} called \textit{R$_{avg}$}, \textit{R$_{min}$}, \textit{S$_{95}$} and \textit{S$_{68}$}.

\textbf{Usability:} Usability with given reliability refers to the extent to which a prediction can be used with uncertainties in decision-making processes. The specific use case of AMs refers to how useable they are in downstream tasks such as path planning. A desirable goal is to predict probability distributions and confidence levels with as little uncertainty as possible while meeting the reliability condition. This leads to sharp residence area predictions with high confidence that the person will be in this area. In path planning, this gives AMs more flexibility for interactions with humans.

\textbf{Real World Applicability:} Considering the challenges of urban traffic and human-machine interaction, specific requirements are necessary for real-world conditions. On the one hand, it must be assumed that only sparse data can be obtained due to occlusions, dynamic observation times, sensor or detection algorithm errors, or outdated mapping data. Consequently, a prediction method must cope with these dynamically occurring challenges. However, if valid reference maps are available, it makes sense to use them. The input side of a model needs to be flexible to cover these circumstances. In HTP, an input size of 3.2 \si{\s} has been established as a de facto standard. Only a few works investigate the behavior using deviating input sizes~\cite{uhlemann}. An initial latency of 3.2 \si{\s} to the first prediction is unsuitable for AV applications. By this time, the critical situation has usually already passed. This circumstance should be given greater consideration. In addition, there are real-time requirements like computational resources, inference- and overall processing time, and scalability in terms of parallel object handling since multiple VRUs can occur simultaneously.

\textbf{Transferability:} Transferability is an essential indicator of how a prediction method works with novel and unseen data. Methods may induce a bias in the data used for training. This can be a bias on the location for generating the data or the general traffic topology and rules in this area. Moreover, many HTP methods use a dataset-specific coordinate system by encoding specific dataset-relevant circumstances. Using an independent human egocentric coordinate system detaches all training data from surrounding positional circumstances. In addition, three or more different data sets from different domains should be used to examine the transferability of a model. Many methods in HTP share the frequent usage of the \textit{ETH/UCY}~\cite{eth}~\cite{ucy} dataset. It consists of five subsets used with a Leave-One-Out (LOO) strategy to evaluate transferability. Schoeller~\cite{scholler} shows correlations between these subsets, making them unsuitable for assessing transferability.

\subsection{Contributions}\label{sec_contributions}
Our contributions are:
\begin{itemize}
  \item We summarize essential requirements, see~\cref{sec_requirements}, a prediction method has to be taken care of to be suitable for real-world autonomous applications. 
  \item In compliance with the aforementioned requirements, we propose an application driven and fast method for probabilistic HTP, providing reliable regions of residence with confidence classifications validated by four datasets to demonstrate our method's performance and robustness.
  \item We demonstrate the importance of reliability evaluation for HTP, giving statistical guarantees that the human resides within the predicted regions with a certain probability to boost robustness and trustworthiness. Therefore, we evaluate \textit{Reliability} and \textit{Sharpness} results of four highly recognized HTP methods.
\end{itemize}
\section{\large Related Work}
\label{sec_related_work}
HTP has been an active research area for many years, resulting in various published methods and architectures to tackle this challenge. Usually, past human tracking information and, if necessary, additional contextual data, like map information or social interactions, are used as inputs. Regression models~\cite{eth} and Multi-Layer-Perceptrons (MLP)~\cite{poly_mlp} are among the first methods to predict future positions. Social-LSTM~\cite{social_lstm} and Social-STGCNN~\cite{social_stgcnn} use unimodal Gaussian distributions to sample discrete trajectory hypotheses but cannot handle multi-modality. Therefore, stochastic prediction models have evolved using bivariate Gaussian Mixtures (GM) based on latent sampled variables, enabled by Conditional Variational-Auto-Encoders (VAE). Trajectron++~\cite{trajectron}, Agentformer~\cite{agent_former}, GroupeNet+\cite{groupnet}, and ExpertTraj~\cite{expert_traj} are some of the ones with the most attention. These methods predict several discrete trajectory hypotheses from the GM without verifying the distribution accuracy and density. Social-GAN~\cite{social_gan} and Sophie~\cite{sophie} use Generative Adversarial Networks (GAN) to handle uncertainty implicitly and to predict discrete future trajectory hypotheses conditioned by random noises. Social-Implicit~\cite{social_implicit} bypasses the GAN discriminator by training an independent one, using Implicit Likelihood Estimation (IMLE) to model the distribution implicitly. MID~\cite{mid} and LED~\cite{led} incorporate denoising probabilistic diffusion to achieve State-Of-The-Art (SOTA) performance at the cost of substantial computational hardware consumption. GATraj~\cite{gatraj} is an attention-based graph model using a Laplacian mixture decoder instead of a GM to mitigate mode collapse. Another approach to predict future trajectories is Flomo\cite{flomo} using analytical computation of normalizing flows~\cite{normalized_flow}. As a further development, FlowChain~\cite{flowchain} uses an expressive stack of Continuously Indexed Flows (CIFs) that allow analytical probability density computation. MotionCNN~\cite{motion_cnn} uses mapping data, renders trajectory data into these maps for trajectory encoding, and uses Convolutional Neural Networks (CNN) for prediction. MemoNet~\cite{memo_net} uses a different approach than many others by imitating the mechanism of retrospective memory in neuropsychology that predicts the movement intentions of agents by looking for similar scenarios in the training data.

The above methods represent the most recognized approaches in HTP. As a downside, the frequent usage of \textit{ETH/UCY} dataset in combination with BoN ADE/FDE evaluation should be questioned. Schoeller~\cite{scholler} investigates the predominantly occurring of vertical and horizontal trajectories, demonstrating the limitations of the dataset. A modified simple CV model achieves comparable ADE/FDE performance, benefitting from the dataset structure. There exist more diverse datasets like \textit{Waymo Motion}~\cite{waymo}, \textit{nuScenes}~\cite{nuscenes}, \textit{inD}~\cite{ind}, \textit{IMPTC}~\cite{imptc}, which have received not enough attention to date for HTP evaluation. Joint-ADE (JADE) and Joint-FDE (JFDE)~\cite{jade} represent two recently introduced meaningful evolutions of ADE/FDE metrics, targeting better evaluation of multi-agent scenarios. Besides FlowChain, Trajectron++, and Social-Implicit, no other research describes and evaluates additional metrics to analyze the quality of the underlying sampling source, which is mainly based on distributions. The real-world usability of most approaches plays a subordinate role, i.e., uncertainty handling, hardware resources, inferencing speed, and scalability.

Some research explicitly targets probabilistic prediction methods to handle uncertainty. This is often achieved by predicting regions, e.g., probability distributions, instead of positions. In~\cite{model-predictive}, an uncertainty estimation created by an unconditional model is presented. This estimation is showcased with a planning method based on predictive control. In~\cite{set_based}, the authors performed set-based predictions using reachability analysis, providing bounded regions that include possible future positions of VRUs, ensuring the safety of planned motions in road traffic. In \cite{zernetsch_holistic}, a primary movement detection classification is used to merge explicit trajectory predictors for different VRU movement types representing an ensemble of unimodal GMs. Despite using single positions, in~\cite{pose}, human body poses are used to predict residence areas. Social-STAGE~\cite{stage} ranks multi-modal Gaussian distributions using probabilities for each mode. The evaluation is limited to \textit{ETH/UCY}, which does not provide multi-modal ground truth. Furthermore, no reliability or Confidence Level (CL) evaluation is provided. TUTR~\cite{tutr} uses a transformer architecture to predict 20 discrete trajectories with associated probability. The number of samples is firmly integrated into the architecture. Moreover, it uses prior knowledge of the dataset to create motion modes used for training and inferencing.

BoN ADE/FDE cannot measure a model's distribution reliability because they do not consider all generated samples. Social-Implicit introduced two metrics to tackle this gap: Average Mahalanobis Distance (AMD) and Average Maximum Eigenvalue (AMV). AMD tries to quantify how close all generated samples are to the ground truth, and AMV describes the overall spread of the predictions. Trajectron++ uses Kernel-Density-Estimated Negativ-Log-Likelihood (KDE-NLL) introduced by~\cite{traj_old}~\cite{thiede} to compare generative methods. It does not evaluate the statistical properties of the underlying distribution: a low NLL does not necessarily mean that a model's distribution is reliable. It only measures how well its density fits the ground truth data. KDE alone can be used to transform discrete samples into Gaussian distributions.~\cite{zernensch_means} and~\cite{zernetsch_holistic} go even further and introduce stochastic methods to evaluate unimodal and arbitrary distributions, called \textit{Reliability} and \textit{Sharpness}. \textit{Reliability} stochastically assesses the correctness of a model's predicted probability densities. \textit{Sharpness} describes a distribution’s expansion regarding CLs related to future prediction time steps. Both metrics stochastically measure the credibility and correctness of distributions and derived CLs, which are essential for uncertainty estimation. This work uses Average- and Minimum Reliability Scores \textit{R$_{avg}$} and \textit{R$_{min}$} derived from~\cite{zernetsch_holistic} to evaluate our model's credibility.
\section{\large Method}
\label{sec_method}
In this section, we present our proposed approach. First, we present the human egocentric coordinate system transformation and describe our model's architecture. Finally, we explain the confidence measurement.

\textbf{Human Egocentric Coordinate Transformation:} To define human trajectories, we use a series of position information $^w\vec{p}_{t}=[^wx_{t}, ^wy_{t}]$ in world coordinates (denoted by $^w$), resulting in a trajectory $^wT_{in;t}=\big\{^w\vec{p}_{t+h}\vert h\in H_{\text{in}}\big\}$ consisting of the humans positions over an input horizon $H_{\text{in}}=\{0,-1,-2,...,-n\}$. We use the future trajectory $^wT_{gt;t}=\big\{^w\vec{p}_{t+h}\vert h\in H_{\text{fc}}\big\}$ with a forecast horizon $H_{\text{fc}}=\{1,2,...,m\}$ as ground truth. Before the trajectories are passed to the model, a transformation of the positions from world to human egocentric coordinates is performed (further referred to as ego coordinates). $\varphi_t$ is the movement direction of the VRU at the current discrete time $t$ and $h\in H_{\text{in}}\cup H_{\text{fc}}$ the relative time offset from $t$. The transformation is performed for both,  $^wT_{in;t}$ and $^wT_{gt;t}$, resulting in $^eT_{in;t}$ and $^eT_{gt;t}$. Ego coordinates solve location-based circumstances, help to avoid a systematic bias, and increase transferability.
 
\begin{figure}[!htb]
	\begin{center}
        \vskip -2mm
		\includegraphics[width = 0.9\columnwidth]{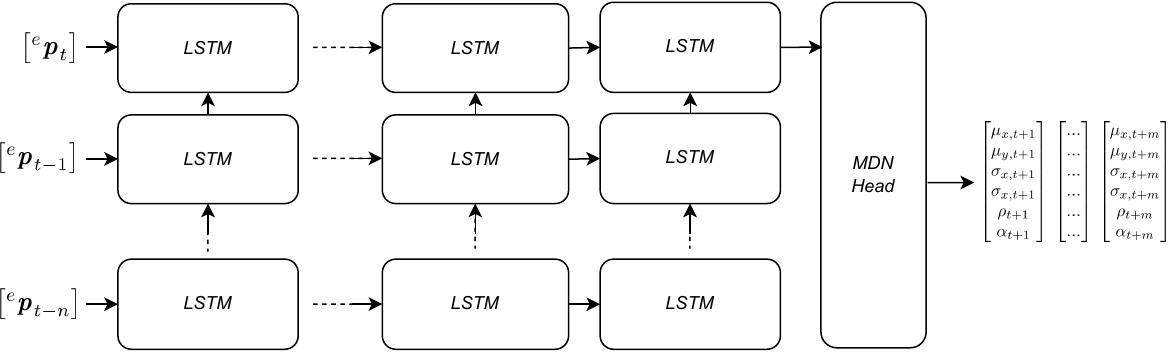}
        \vskip -1mm
		\caption{Network architecture. We use a stacked LSTM for dynamic input horizon handling and a bivariate Mixture Density Head for the probabilistic predictions.}
		\label{fig:mdn_arch}
	\end{center}
	\vskip -10mm
\end{figure}

\textbf{Model Architecture:} We use a stacked Long-Short-Term-Memory (LSTM) to cope with dynamic input horizons and a Mixture Density Network (MDN)~\cite{mdn_bishop} as head, illustrated by~\cref{fig:mdn_arch}. In \cite{zernensch_means}, it was discovered that a single normal distribution per forecast horizon is not sufficient to model the predicted uncertainty reliably. In the MDN Head, we model the uncertainty multi-modally and semi-supervised by MDNs, resulting in a mixture of bivariate Gaussian distributions. As activation functions for parameters of the Gaussian components $\mathcal{F}_{m;t+h}$, we use $\sigma(\vec{o_{\sigma}})=exp(\vec{o_{\sigma}}) + 1 + \epsilon_{\sigma}$ and $\rho(\vec{o_{\sigma}})=tanh(\vec{o_{\sigma}}) \cdot \epsilon_{\rho}$ with $\sigma$ greater than zero, and $\rho$ within $\left]-1,1\right[$, with $o_{\sigma}$ being $\sigma_{x}$ and $\sigma_{y}$~\cite{zernensch_means}. For the weights of the components $c_{m:t+h}$, we use soft-max activation.

The mixtures for each forecasted time horizon $h \in H_{\text{fc}}$ are built by~\cref{equ:mixture_dist}. The joint distribution $\mathcal{D}_{t+h}$ consists of the predicted likelihood $\mathcal{F}_{m;t+h}$ of a mode $m$, weighted by a probability $c_{m;t+h}$ for the respective mode. As loss function,~\cref{equ:NLL}, we use the Negative Log-Likelihood (NLL) over the mixture distribution and average the loss over each forecast horizon $h \in H_{\text{fc}}$.

\begin{center}
\vskip -11mm
\begin{tabular}{p{6cm}p{6cm}}
  \begin{equation}\label{equ:mixture_dist}
    \resizebox{0.85\hsize}{!}{
        $\mathcal{D}_{t+h}(\vec{p}_{t+h}) = \sum_{m \in M} c_{m:t+h}\cdot \mathcal{F}_{m;t+h}(\vec{p}_{t+h})$
        }
  \end{equation}
  &
  \begin{equation}\label{equ:NLL}
      \resizebox{0.75\hsize}{!}{
        $L_{NLL} = -\sum_{h \in H_{\text{fc}}} \log \mathcal{D}_{t+h}(\vec{p}_{t+h})$
        }
  \end{equation}
\end{tabular}
\end{center}
\vskip -8mm

\textbf{Confidence Measures:} Path planning and interaction handling tasks depend on uncertainty estimations. Therefore, we use confidence measurements. For a given distribution, we define a confidence set $\Omega(1-\alpha)$ as a set of points with the CL of $1-\alpha$, with $\alpha\in\left[0.0,1.0\right]$. To obtain the CL $1-\alpha(\vec{p})$ of a point $\vec{p}$, we search for points $\vec{z}\in\mathbb{R}^2$, which have a probability density level greater than or equal to the density $\mathcal{D}(\vec{p})$~\cite{zernetsch_holistic}. Doing this for every point in $\mathbb{R}^2$ is not feasible in practice, so we draw $N$ random samples $\mathbf{Z}\sim\mathcal{D}$ from the distribution $\mathcal{D}$ to find these points. The CL estimate $1-\alpha(\vec{p})$ of the observation $\vec{p}$ is determined by:

\begin{gather1}[0.9]
    1-\alpha(\vec{p}) = \frac{1}{N}\sum_{\vec{z}\in \mathbf{Z}}
    \begin{cases}
        1, & \text{ if } \mathcal{D}(\vec{z}) \geq \mathcal{D}(\vec{p}) \\
        0,              & \text{otherwise}
    \end{cases}
\end{gather1}

\begin{figure}[h]
	\begin{center}
		\vskip -6mm
		\includegraphics[width = 0.5\columnwidth]{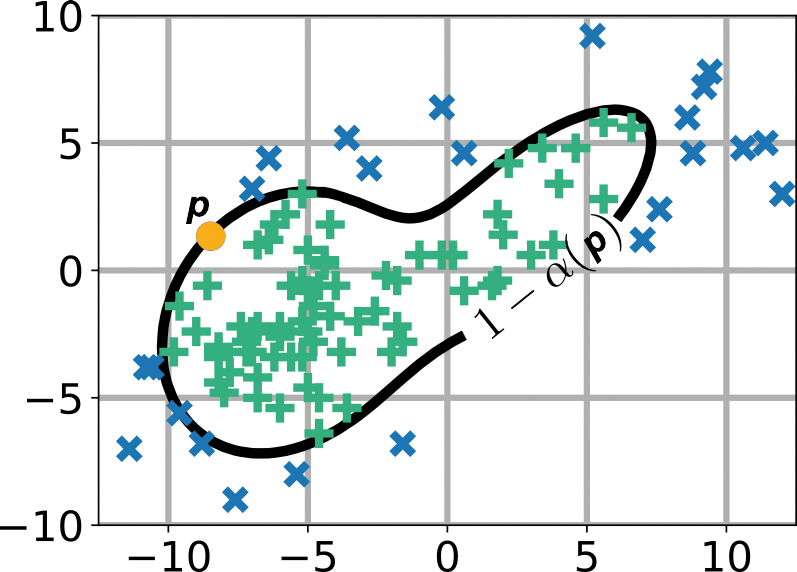}
		\vskip -2mm
		\caption{Confidence level $1-\alpha(\vec{p})$ (black curve) of $\vec{p}$ (orange dot), with $N$ random samples $\vec{z}\in \mathbf{Z}$ with $\mathcal{D}(\vec{z})\geq\mathcal{D}(\vec{p})$ as green, and $\mathcal{D}(\vec{z})<\mathcal{D}(\vec{p})$ as blue markers.}
		\label{fig:distribution_eval}
	\end{center}
	\vskip -6mm
\end{figure}

~\cref{fig:distribution_eval} illustrates the CL estimation using an exemplary distribution. By knowing the CL of arbitrary points, we can build confidence sets $\Omega(1-\alpha)$ as a set of points with at least the CL $1-\alpha$:

\begin{equation}
    \resizebox{0.50\hsize}{!}{
       $\Omega(1-\alpha)=\big\{\vec{p}\vert\vec{p}\in\mathbb{R}^2\land\overbrace{1-\alpha(\vec{p})}^{\text{confidence level of }\vec{p}} \geq 1-\alpha\big\}$
    }
\end{equation}

Confidence sets represent future residence areas with a certain probability. Knowing this area is mandatory in downstream tasks like path planning to ensure safe interaction with humans. CLs enable the estimate of the probability of a collision if the planned path of an autonomous system is given. 

\section{\large Experiments}
\label{sec_dataset}

\subsection{Experimental Setup}\label{sec_setup}
\textbf{Datasets:} Experiments were conducted regarding the in Chapter~\ref{sec_requirements} described requirements. Therefore, we use four traffic-related datasets recorded by different observation techniques and circumstances to demonstrate our overall objective. The composition consists of the vehicle-based datasets \textit{Waymo Motion}, \textit{nuScenes}, the drone-based set \textit{inD}, and the infrastructure-based \textit{IMPTC} datasets. The \textit{ETH/UCY} dataset is not part of this composition; it is used later to compare our method with current SOTAs and to demonstrate the importance of \textit{Reliability} and \textit{Sharpness} metrics.

\textit{The Waymo Motion} dataset consists of over 100 K scenes collected by a sensor-equipped vehicle in public traffic. Each scene includes multiple road users and types. \textit{nuScenes} consists of 1 K driving scenes from Boston and Singapore, including multiple road users. \textit{inD} uses drones to monitor four public intersections, consists of 33 recording sessions, and contains various moving VRUs and interactions. \textit{IMPTC} uses a camera- and LiDAR-equipped public intersection infrastructure for data recording. It consists of 270 sequences and provides a complete environmental perception of all traffic participants. It includes additional context information like traffic light signal status, weather data, and mapping information. To bypass different sampling rates, we applied Cubic Spline to resample all datasets at a fixed rate of 10.0 Hz. ~\cref{table:datasets_overview} compares the different datasets' key facts. We conduct multiple evaluations using all four presented datasets for transferability, scalability, and real-world applicability. 

The \textit{ETH/UCY} dataset uses a fixed-mounted and elevated camera to monitor different public pedestrian places like the pedestrian walk and entrance area of a hotel, shopping mall, and university. It consists of five locations and only provides pedestrian trajectories. We define $H_{\text{in}}$ and $H_{\text{fc}}$ for training and experiments on all datasets to fit the commonly used procedure of 3.2 \si{\s} input- and 4.8 \si{\s} output horizon with an untouched sample rate of 2.5 Hz.

\begin{table}[htb]
\vskip -6mm
\centering
\caption{Key facts and Comparison of all used datasets.}
\vskip -2mm
\label{table:datasets_overview}
\resizebox{0.8\columnwidth}{!}
{    
    \begin{tabular}{c||c|c|c|c||c}
      \toprule
      \toprule
      \textbf{Categories} & \textbf{nuScenes} & \textbf{Waymo} & \textbf{inD} & \textbf{IMPTC} & \textbf{ETH/UCY} \\
      \midrule
      \textit{Type} & vehicle & vehicle & drone & infra & infra \\
      \textit{Sensors} & \makecell[c]{camera\\lidar, radar} & \makecell[c]{camera\\lidar} & camera & \makecell[c]{camera\\lidar} & camera \\
      \textit{\makecell[c]{Original\\Sample Rate}} & 2.0 Hz & 10.0 Hz & 25.0 Hz & 25.0 Hz & 2.5 Hz  \\
      \textit{\makecell[c]{Modified\\Sample Rate}} & 10.0 Hz & 10.0 Hz & 10.0 Hz & 10.0 Hz & 2.5 Hz  \\
      \textit{Location} & road traffic & road traffic & road traffic & road traffic & public places \\
      \bottomrule
      \multicolumn{6}{l}{}\\
    \end{tabular}  
}
\vskip -8mm
\end{table}

We only use pedestrian tracks with a length equal to or greater than the combined input and output horizon for training and evaluation. Longer tracks are split into subsamples. By default, \textit{IMPTC}, \textit{Waymo}, and \textit{nuScenes} provide train, train-eval, and test splits. For \textit{inD}, we used all recordings from intersections two and three for training, intersection four for train evaluation, and intersection one for testing. For \textit{ETH/UCY}, we used the established standard LOO splits introduced by Social-GAN~\cite{social_gan}.

\subsection{Evaluation Methods}\label{sec_eval_methods}
\textbf{Reliability \& Sharpness:} We can evaluate the reliability of the prediction framework by calculating the estimated confidence level $1-\alpha(\vec{p})$ for every corresponding pair of predicted distribution and ground truth point. The observed relative frequency of occurrences $f_o(1-\alpha)$ should match $1-\alpha$. We visually evaluate the reliability calibration by plotting the observed frequency over $1-\alpha$, ideally resulting in a line with slope one. As comparative metrics we build $f_{o;t+h}(1-\alpha)$ from every corresponding pair of predicted distribution $\mathcal{D}_{t+h}$ and ground truth position $\vec{p}_{t+h}$ in the validation set and search for the largest (~\cref{equ:murphy_rel_score}) and average (~\cref{equ:mean_rel_score}) distance from the ideal value $1-\alpha$ over all $\alpha \in A=\{0.01,0.02,...0.99\}$ and all time horizons $h \in H_{\text{fc}}$:

\begin{center}
\vskip -8mm
\begin{equation}\label{equ:murphy_rel_score}
    \resizebox{0.50\hsize}{!}{
    $\hat{\Gamma} = 1 - \max_{h \in H_{\text{fc}}}\max_{\alpha \in A} |1-\alpha-f_{o;t+h}(1-\alpha)|$
    }
\end{equation} \\
\vskip -8mm
\begin{equation}\label{equ:mean_rel_score}
    \resizebox{0.60\hsize}{!}{
     $\bar{\Gamma} = 1 - \frac{1}{|H_{\text{fc}}| \cdot |A|}\sum_{h \in H_{\text{fc}}}\sum_{\alpha \in A} |1-\alpha-f_{o;t+h}(1-\alpha)|$
     }
\end{equation}
\end{center}

$\bar{\Gamma}$ represents the \textit{Average Reliability Score} \textit{R$_{avg}$} and $\hat{\Gamma}$ is the \textit{Minimum Reliability Score} \textit{R$_{min}$}. For these metrics, the perfect score is 1, representing a perfectly reliable HTP system, and the worst score is 0. A thoroughly reliable system has an \textit{Average Reliability} $\bar{\Gamma}$ of $\geqslant$ >= 0.95 and a \textit{Minimum reliability} score $\hat{\Gamma}$ of $\geqslant$ >= 0.90. For a certain confidence level $1-\alpha$ of a distribution, we define the \textit{Sharpness} $\kappa(1-\alpha)$ in $\si{\meter^2\per\second}$ as the volumetric measure of the confidence set $\Omega(1-\alpha)$. We use the first and second standard deviation, i.e., 95 \% and 68 \% quantiles, to report specific \textit{Sharpness} levels called \textit{S$_{95}$} and \textit{S$_{68}$}.

\textbf{Average- and Final-Displacement Error:} ADE is the average L2 distance between the predicted trajectory $T$ and ground truth trajectory $T_{gt}$ over time, represented by a number of discrete future time steps $m$, and FDE is the L2 distance between the endpoints of the predicted and ground truth trajectories.

\textbf{Implementation Details:} Our LSTM has 8 Hidden Layers, and the MDN Head is defined to have a fixed output size of six parameters per forecast horizon $H_{\text{fc}}$. Architectural parameters have been determined by grid search. For multi-modality, we set the number of Gaussians to three, which is reasonable for human motion~\cite{diss_kress}. We used ADAM Optimizer with an initial learning rate of 10$^{-3}$, a linear regression towards 10$^{-7}$, a batch size of 1024, and trained our model for 2500 epochs. The training data will randomly be re-ordered for every epoch. We used an NVIDIA RTX 3090 GPU with an AMD Ryzen 5900X processor for training and evaluation. Different computing platforms are evaluated for inference- and computing time measurements, summarized in~\cref{tab:computing_plattforms}.

\subsection{Results}\label{sec_results}
\begin{figure}[htb!]
    \begin{center}
        \vskip -6mm
        \includegraphics[width = 1.0\columnwidth]{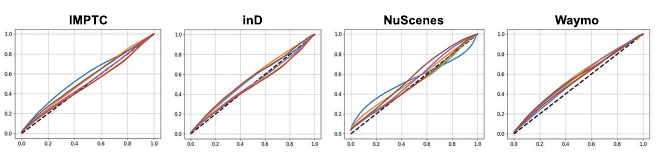}
        \vskip -2mm
        \caption{Calibration plots for reliability and usability check. Curves for predictions at ${t+0.8s}$ (blue), at ${t+1.6s}$ (orange), at ${t+2.4s}$ (green), at ${t+3.2s}$ (red), at ${t+4.0s}$ (purple), at ${t+4.8s}$ (brown) and ideal behavior (black dashed line). The horizontal axis is the expected CL, and the vertical axis is the observed frequency of CLs.}
        \label{fig:overall_reliability_plots}
    \end{center}
    \vskip -6mm
\end{figure}

\begin{figure}[h!]
	\begin{center}
		\vskip -2mm
		\includegraphics[width = 1.0\columnwidth]{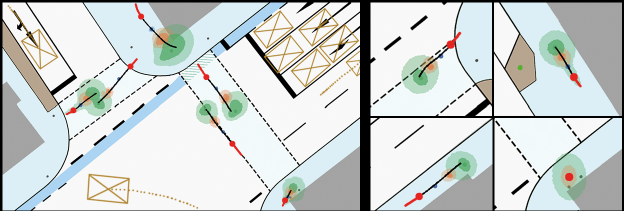}
		\vskip -2mm
		\caption{Example results for the \textit{IMPTC} dataset. Past input trajectory (red), ground truth (black line), predicted CLs at $t{+1s}$ (blue area), $t{+2s}$ (orange area), and $t{+3s}$ (green area). Inner contours represent the 68 \%, and outer contours the 95 \% CL.}
		\label{fig:imptc_world_plots}
	\end{center}
	\vskip -6mm
\end{figure}

\textbf{Reliability and Usability:} We report our results utilizing the metrics described in~\cref{sec_eval_methods} in~\cref{table:multi_scores}. First, we test our model's performance in terms of reliability and usability. Therefore, we trained the model using the train data from all four datasets and tested it with the test splits of all datasets individually (Multi-Train and Single-Test). Our method works reliably on all four datasets, achieving an \textit{R$_{avg}$} of 95.6 \% and an \textit{R$_{min}$} of 88.0 \%. The minimum scores are slightly below 90 \%, which leaves room for improvement. ~\cref{fig:overall_reliability_plots} visualizes the corresponding reliability calibration plots. The curves for selected forecast horizons come close to the ideal line. Looking at the \textit{Shaprness}, the average values of \textit{S$_{68}$} 0.4 ${m^2/s}$ and \textit{S$_{95}$} 1.9 ${m^2/s}$, which is a reasonable time-related scaling behavior of expanding confidence areas over the forecast horizon for typical human movements. ADE and FDE results of 0.28 \si{\m} and 0.55 \si{\m} are consistently good across all datasets regarding the forecast horizon of up to 4.8 \si{\s}. Overall, our method demonstrates a robust and reliable performance across multiple datasets. Downstream tasks can trust the model's human trajectory predictions.~\cref{fig:imptc_world_plots} illustrates some example predictions for the \textit{IMPTC} dataset.

\begin{table}[h!]
\vskip -6mm
\centering
\caption{Evaluation results for the Multi-Train and Single-Test evaluation.}
\vskip -2mm
\label{table:multi_scores}
\resizebox{0.65\columnwidth}{!}
{    
    \begin{tabular}{c||c|c|c|c||c}
      \toprule
      \toprule
      \textbf{Scores} & \textbf{IMPTC} & \textbf{inD} & \textbf{nuScenes} & \textbf{Waymo} & \textbf{Avg.}\\
      \midrule
      R$_{avg}$ ${(\%)}$ & 96.3 & 96.2 & 95.3 & 94.6 & \textbf{95.6} \\
      R$_{min}$ ${(\%)}$ & 87.9 & 90.9 & 84.4 & 88.8 & \textbf{88.0}  \\
      S$_{68}$ ${(m^2/s)}$ & 0.4 & 0.3 & 0.4 & 0.5 & \textbf{0.4} \\
      S$_{95}$ ${(m^2/s)}$ & 2.0 & 1.6 & 2.1 & 1.8 &  \textbf{1.9} \\
      $_{min}$ADE$_{20}$ ${(m)}$ & 0.30 & 0.27 & 0.27 & 0.26 & \textbf{0.28} \\
      $_{min}$FDE$_{20}$ ${(m)}$ & 0.63 & 0.53 & 0.50 & 0.52 & \textbf{0.55} \\
      \bottomrule
    \end{tabular}   
}
\vskip -6mm
\end{table}

\textbf{Transferability:}
To evaluate transferability, we trained the model using the train data from every dataset individually and tested it with the test splits of each dataset individually for cross-validation. The individual results in~\cref{table:cross_scores} demonstrate that our method can generalize well across multiple datasets, keeping \textit{Reliability} levels, indicating robustness against domain changes. Only the individual \textit{IMPTC}-trained model has some struggle when being applied to \textit{inD} and \textit{nuScenes} in terms of \textit{R$_{min}$} of 71.2 \% and 70.2 \%. Single forecast horizon steps create outliers, indicating that the training has stopped too early. 

\begin{table}[htb]
\vskip -8mm
\footnotesize
\centering
\setlength{\tabcolsep}{1mm}{\caption{\small Cross-validation results for transferability test. Results are structured as follows: first row: R$_{avg}$/R$_{min}$ in \textit{${(\%)}$}, second row: S$_{68}$/S$_{95}$ in ${(m^2/s)}$, and third row $_{min}$ADE$_{20}$/$_{min}$FDE$_{20}$ in \textit{${(m)}$}.
\label{table:cross_scores}}
}
\vskip -2mm
\fontsize{8.8}{9.8}\selectfont
\resizebox{0.75\columnwidth}{!}
{
    \begin{tabular}{c||c|c|c|c||c}
    \hline
    \hline       
        \makecell{Train Set} 
        &\makecell{IMPTC}
        &\makecell{inD}
        &\makecell{nuScenes}
        &\makecell{Waymo}
        &\makecell{\textbf{Avg}}
        \\
    \hline
        IMPTC & \makecell[c]{96.5/92.7\\0.6/1.9\\0.27/0.58} & \makecell[c]{93.3/71.2\\0.6/1.9\\0.33/0.69} & \makecell[c]{93.5/70.2\\0.9/2.5\\0.35/0.66} & \makecell[c]{95.2/90.5\\0.6/2.1\\0.28/0.60} & \textbf{\makecell[c]{94.6/81.2\\0.7/2.1\\0.31/0.63}} \\
            \hline
        inD & \makecell[c]{96.9/89.3\\0.2/1.3\\0.30/0.64} & \makecell[c]{98.2/95.9\\0.2/1.2\\0.31/0.65} & \makecell[c]{96.2/86.5\\0.3/1.8\\0.29/0.56} & \makecell[c]{94.8/91.1\\0.3/1.2\\0.32/0.59} & \textbf{\makecell[c]{96.5/90.7\\0.25/1.4\\0.31/0.61}} \\
            \hline
        nuScenes & \makecell[c]{96.8/90.6\\0.5/2.0\\0.33/0.69} & \makecell[c]{96.8/87.1\\0.3/1.4\\0.35/0.72} & \makecell[c]{97.1/91.9\\0.6/2.0\\0.28/0.55} & \makecell[c]{92.6/88.4\\0.5/1.7\\0.39/0.67} & \textbf{\makecell[c]{95.8/89.5\\0.3/1.8\\0.34/0.66}} \\
            \hline
        Waymo & \makecell[c]{97.2/92.3\\0.4/1.7\\0.33/0.68} & \makecell[c]{97.5/90.0\\0.3/1.6\\0.36/0.73} & \makecell[c]{96.8/88.9\\0.3/1.7\\0.29/0.56} & \makecell[c]{96.5/93.5\\0.3/1.4\\0.24/0.49} & \textbf{\makecell[c]{97.0/91.2\\0.3/1.6\\0.31/0.62}} \\
    \hline
    \end{tabular}
    }
    \vskip -4mm
\end{table}

\textbf{Real-World Applicability:} In autonomous systems, the observation horizon can vary. Therefore, we evaluate the impact of the observation horizon on prediction quality and reliability. As shown in~\cref{table:input_horizon_scaling}, we use different observation horizons to evaluate our models' performance. It copes well with shorter observation horizons, which are typically given under actual traffic conditions. It should be emphasized that a reliable prediction is possible even for short observation horizons. However, further observations improve reliability and sharpness. The same applies to the ADE and FDE metrics. Overall, an input horizon of 0.5 \si{\s} is sufficient for our method to achieve near-full performance. In~\cite{uhlemann}, Trajectron++, Agentformer, Social-Implicit, and YNet~\cite{ynet} were examined for their ability to cope with shorter observation horizons analyzing ADE and FDE metrics. In conclusion, only Trajektron++ can cope with dynamic input lengths; all other tested methods demonstrate increasing ADE and FDE results of up to 3x degradation. Only Trajectron++ uses additional velocity values, besides positional information, as we do, as model input. We also conducted the input horizon scaling test without velocity values, which resulted in highly reduced results. Therefore, velocity is an essential input feature to boost short observation times performance.

\begin{table}[h!]
\vskip -2mm
\centering
\caption{Input horizon scaling tested with different input lengths.}
\vskip -2mm
\label{table:input_horizon_scaling}
\resizebox{0.8\columnwidth}{!}
{
    \begin{tabular}{c||c|c|c|c|c|c|c}
    \hline
    \hline
    \textbf{Dataset} 
    & \textbf{\makecell[c]{$H_{\text{in}}$\\${(ms)}$}} 
    & \textbf{\makecell[c]{R$_{avg}$\\${(\%)}$}} 
    & \textbf{\makecell[c]{R$_{min}$\\${(\%)}$}} 
    & \textbf{\makecell[c]{S$_{68}$\\${(m^2/s)}$}} 
    & \textbf{\makecell[c]{S$_{95}$\\${(m^2/s)}$}} 
    & \textbf{\makecell[c]{$_{min}$ADE$_{20}$\\${(m)}$}} 
    & \textbf{\makecell[c]{$_{min}$FDE$_{20}$\\${(m)}$}}\\
    \hline
    \multirow{4}{*}{\begin{turn}{270}\textbf{Multi}\end{turn}} 
                                   & 100 & 86.7 & 68.4 & 0.9 & 3.7 & 0.42 & 0.85 \\
                                   & 500 & 92.9 & 84.3 & 0.8 & 2.9 & 0.38 & 0.65 \\
                                   & 1000 & 94.3 & 86.3 & 0.6 & 2.3 & 0.30 & 0.59 \\
                                   & 2000 & 95.2 & 87.5 & 0.5 & 2.2 & 0.28 & 0.56 \\
    \hline
    \end{tabular}
    }
\vskip -6mm
\end{table}

\textbf{Inferencing:} We furthermore evaluate the runtime of our model on different platforms to demonstrate its portability, speed, and minor hardware requirements using a batch size of 128.~\cref{tab:computing_plattforms} illustrates our method's inferencing and post-processing performance, achieving excellent results on all platforms, with a total runtime of only 2.4 ms for 128 parallel predictions on a 10 W embedded Nvidia Tegra TX2 platform. Using a GPU speeds up the inferencing by 2-4x, but having one is optional. Our model is the fastest compared to other approaches from~\cref{table:eth_ade_comparison}. From a resource and inferencing perspective, our approach is suitable for usage within AVs.

\begin{table}[!htb]
\vskip -4mm
\centering
\caption{Inferencing comparison between different computing platforms w/wo GPU. All measurements use a batch size of 128.}
\vskip -2mm
\label{tab:computing_plattforms}
\resizebox{1.0\columnwidth}{!}
{    
    \begin{tabular}{c|c|c|c|c}
    \hline
    \hline
    \textbf{Target} & \textbf{Plattform} & \textbf{Inference Time} & \textbf{Post Processing Time} &\textbf{Total Time}\\
    \hline
    \multirow{3}{*}{\begin{turn}{270}\textbf{GPU}\end{turn}}
       & Desktop~$^{1}$  ${(x86)}$ & 0.3 ms & 0.4 ms & 0.7 ms \\
       & Laptop~$^{2}$  ${(x86)}$) & 0.5 ms & 0.3 ms & 0.8 ms \\
       & Embedded~$^{3}$  ${(ARM)}$ & 1.2 ms & 1.2 ms & 2.4 ms \\
    \hline
    \multirow{3}{*}{\begin{turn}{270}\textbf{CPU}\end{turn}}
       & Desktop~$^{1}$  ${(x86)}$ & 1.2 ms & 0.4 ms & 1.6 ms \\
       & Laptop~$^{2}$  ${(x86)}$ & 1.1 ms & 0.3 ms & 1.4 ms  \\
       & Embedded~$^{3}$  ${(ARM)}$ & 2.8 ms & 1.2 ms & 4.0 ms  \\
    \hline
    \multicolumn{5}{l}{$^{1}$\footnotesize{(AMD Ryzen 3700X + Nvidia RTX 3060)}, $^{2}$\footnotesize{(AMD Ryzen 7840U + Radeon 780m)}, $^{3}$\footnotesize{(Nvidia Jetson Tegra TX2)}} \\
    \end{tabular} 
}
\vskip -6mm
\end{table}

\begin{figure}[htb!]
    \begin{center}
        \vskip -2mm
        \includegraphics[width = 1.0\columnwidth]{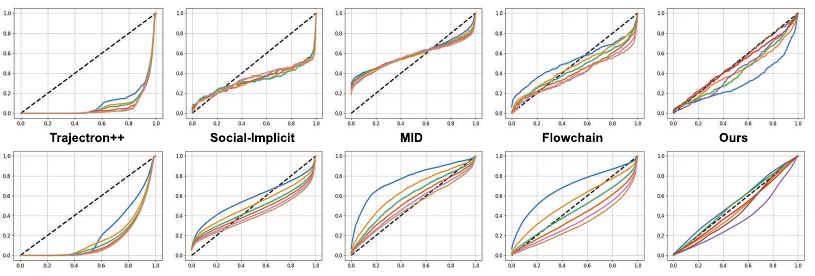}
        \vskip -2mm
        \caption{Reliability calibration plots for method comparison using ETH/UCY dataset. Colors and axis are equal to~\cref{fig:overall_reliability_plots}. Upper row show \textit{ETH} subtest, lower row is \textit{Univ} subtest.}
        \label{fig:eth_reliability_plots}
    \end{center}
    \vskip -8mm
\end{figure}

\textbf{Comparison with Others:} 
To compare our work with others, we select four highly recommended methods. We take Trajectron++, one of the most influential works of the last couple of years in HTP. Parts of it are used in multiple subsequent works like \cite{mid} or \cite{flowchain}. Trajectron represents the class of CVAEs. Furthermore, we selected Social-Implicit to represent GAN architectures and their effort to demonstrate the weaknesses of ADE/FDE and NLL for distribution evaluation. MID is the first diffusion-based method, representing Transformer architectures, and FlowChain represents flow-based approaches, including the additional effort to evaluate distribution densities. We implemented the \textit{Reliability} and \textit{Sharpness} evaluation into named frameworks to measure their performance besides ADE/FDE. We thank the authors for providing their code and documentation. The \textit{ETH-UCY} dataset serves as a link between the four selected methods and our work. We trained our model the same way using the LOO strategy. The results are provided by~\cref{table:eth_reliability_comparison}, listing \textit{Reliability} and ADE/FDE scores next to each other. The corresponding calibration plots for \textit{ETH} and \textit{Univ} subtests are provided in~\cref{fig:eth_reliability_plots}. None of the tested models can achieve suitable \textit{Reliability} scores. Our model offers the best scores with room for improvement. It must be noted that ADE/FDE performance does not directly translate into good \textit{Reliability} results.

\begin{table}[htb]
\vskip -2mm
\footnotesize
\centering
\setlength{\tabcolsep}{1mm}{\caption{\small Comparison with SOTA on \textit{ETH/UCY} dataset. Results are structured as follow: first row: R$_{avg}$/R$_{min}$ in \textit{${(\%)}$} and second row $_{min}$ADE$_{20}$/$_{min}$FDE$_{20}$ in \textit{${(m)}$}. The best is in \textbf{bold}, and the second is \underline{underlined}.
\label{table:eth_reliability_comparison}}
}
\vskip -2mm
\fontsize{10.8}{11.8}\selectfont
\resizebox{1.0\columnwidth}{!}
{
    \begin{tabular}{c||c|c|c|c||c}
    \hline
    \hline       
        \multirow{1}{*}{\textbf{Subset}} 
        &\makecell[c]{\textbf{Trajectron++$^{1}$}}
        &\makecell[c]{\textbf{Social Implicit}}
        &\makecell[c]{\textbf{$~$MID$^{1}$}}
        &\makecell[c]{\textbf{$~$FlowChain}}
        &    \multirow{1}{*}{\textbf{Ours}}
        \\
    \hline
        ETH  & \makecell[c]{58.3/21.4\\0.61/1.02} & \makecell[c]{82.0/56.6\\0.66/1.44} & \makecell[c]{89.9/75.6\\0.39/0.66} & \makecell[c]{85.7/67.6\\0.55/0.99} & \makecell[c]{91.9/67.4\\0.51/0.92} \\
            \hline
        Hotel & \makecell[c]{64.5/29.5\\0.19/0.28} & \makecell[c]{70.1/52.0\\0.20/0.36} & \makecell[c]{89.0/55.1\\0.13/0.22} & \makecell[c]{85.7/67.6\\0.20/0.35} & \makecell[c]{90.7/78.1\\0.16/0.27} \\
            \hline
        Univ & \makecell[c]{66.8/42.4\\0.30/0.54} & \makecell[c]{89.3/75.4\\0.31/0.6} & \makecell[c]{84.3/73.8\\0.22/0.45} & \makecell[c]{87.5/69.0\\0.29/0.54} & \makecell[c]{94.7/77.8\\0.26/0.51} \\
            \hline
        Zara1 & \makecell[c]{61.1/21.8\\0.24/0.42} & \makecell[c]{90.7/74.0\\0.25/0.50} & \makecell[c]{81.3/69.5\\0.17/0.30} & \makecell[c]{89.0/55.1\\0.22/0.40} & \makecell[c]{89.4/72.8\\0.23/0.47} \\
            \hline
        Zara2 & \makecell[c]{73.9/39.4\\0.18/0.51} & \makecell[c]{81.3/54.3\\0.22/0.35} & \makecell[c]{78.7/55.3\\0.13/0.27} & \makecell[c]{80.3/47.8\\0.20/0.34} & \makecell[c]{88.6/66.1\\0.16/0.34} \\
    \hline
    \hline
        \textbf{Avg.} & \makecell[c]{64.9/30.9\\0.30/0.51} & \makecell[c]{82.7/62.5\\0.33/0.67} & \makecell[c]{84.6/\underline{65.9}\\\textbf{0.21}/\textbf{0.38}} & \makecell[c]{\underline{86.5}/62.2\\0.29/0.52} & \makecell[c]{\textbf{91.1}/\textbf{72.5}\\\underline{0.26}/\underline{0.50}} \\
    \hline
    \multicolumn{6}{l}{$^{1}$\footnotesize{Updated results according to implementation issue \#53 on the Trajectron++ GitHub page}} \\ 
    \end{tabular}
    }
    \vskip -8mm
\end{table}

In~\cref{table:eth_ade_comparison}, we compare our approach against the current SOTA solely related to ADE/FDE using the \textit{ETH-UCY} dataset. Our model can not compete with the current SOTA from an absolute performance point. Still, it can exceed models like Trajectron++ or Flowchain, which is remarkable due to its non-use of additional context information like social interaction or map data. Compared to Social-LSTM, the yet best LSTM-based model, we can improve ADE and FDE performance by 42 \%, respectively, 30 \%. There are no comparable approaches to date in evaluating distributions and reliability. We added the modified CV model from Scholler~\cite{scholler} to demonstrate how well a relatively simple model performs compared to the current SOTA when using this dataset. As explained in~\cref{sec_related_work}, the dataset has some weaknesses, benefitting simple approaches solely focusing on ADE/FDE evaluation. This consideration changes significantly when taking \textit{Reliability} into account. As shown in~\cref{table:eth_reliability_comparison}, good ADE/FDE results do not automatically lead to good \textit{Reliability}.

\begin{table}[htb]
\vskip -6mm
\footnotesize
\centering
\setlength{\tabcolsep}{1mm}{\caption{\small Comparison with state-of-the-art and two legacy models on \textit{ETH-UCY} dataset. The results are structured as follows: $_{min}$ADE$_{20}$/ $_{min}$FDE$_{20}$ \textit{${(m)}$}.
\label{table:eth_ade_comparison}}
}
\vskip -2mm
\fontsize{10.8}{11.8}\selectfont
\resizebox{\columnwidth}{!}
{
    \begin{tabular}{c||cccccccc||c}
    \hline
    \hline       
        \multirow{2}{*}{Subset} 
        &\makecell[c]{CVM\\\cite{scholler}}
        &\makecell[c]{Social-LSTM\\\cite{social_lstm}}
        &\makecell[c]{Trajectron++$^{1}$\\\cite{trajectron}}
        &\makecell[c]{Social\\Implicit\\\cite{social_implicit}}
        &\makecell[c]{$~$MID$^{1}$\\\cite{mid}$~$}
        &\makecell[c]{$~$GATraj\\\cite{gatraj}$~$}
        &\makecell[c]{$~$LED\\\cite{led}$~$}
        &\makecell[c]{$~$FlowChain\\\cite{flowchain}$~$}
        &    \multirow{2}{*}{\textbf{Ours}}
        \\
        & \color{blue}{\scriptsize{RA-L'20}}
        & \color{blue}{\scriptsize{CVPR'16}} 
        & \color{blue}{\scriptsize{ECCV'20}} 
        & \color{blue}{\scriptsize{ECCV'22}} 
        & \color{blue}{\scriptsize{CVPR'22}}
        & \color{blue}{\scriptsize{ISPRS'23}} 
        & \color{blue}{\scriptsize{CVPR'23}} 
        & \color{blue}{\scriptsize{ICCV'23}} 
        &
        \\
    \hline
        ETH & 0.43/0.80 & 0.73/1.09 & 0.61/1.02 & 0.66/1.44 & \underline{0.39}/0.66 & \textbf{0.26/0.42} & \underline{0.39}/\underline{0.58} & 0.55/0.99 & 0.51/0.92 \\
        Hotel & 0.19/0.35 & 0.49/0.79 & 0.19/0.28 & 0.20/0.36 & 0.13/0.22 & \textbf{0.10/0.15} & \underline{0.11}/\underline{0.17} & 0.20/0.35 & 0.16/0.27 \\
        Univ & 0.34/0.71 & 0.41/0.67 & 0.30/0.54  & 0.31/0.60 & \underline{0.22}/0.45 & \textbf{0.21/0.38} & 0.26/\underline{0.43} & 0.26/0.51 & 0.24/0.53 \\
        Zara1 & 0.24/0.48 & 0.27/0.47 & 0.24/0.42 & 0.25/0.50 & 0.17/0.30 & \textbf{0.16}/\underline{0.28} & \underline{0.18}/\textbf{0.26} & 0.22/0.4 & 0.23/0.47 \\
        Zara2 & 0.21/0.45 & 0.33/0.56 & 0.18/0.32 & 0.22/0.43 & \underline{0.13}/0.27 & \textbf{0.12/0.21} & \underline{0.13}/\underline{0.22} & 0.20/0.34 & 0.16/0.34 \\
    \hline
        \textbf{Avg.} & 0.28/0.56 & 0.45/0.72 & 0.30/0.51 & 0.33/0.67 & \underline{0.21}/0.38 & \textbf{0.17/0.29} & \underline{0.21}/\underline{0.33} & 0.29/0.52 & 0.26/0.50 \\
    \hline
    \hline
     \makecell[c]{Distribution\\Evaluation}& \ding{53} & \ding{53} & \ding{51} & \ding{53} & \ding{53} & \ding{53} & \ding{53} & \ding{51} & \ding{51} \\
    \hline
     \makecell[c]{Reliability\\Calibration} & \ding{53} & \ding{53} & \ding{53} & \ding{53} & \ding{53} & \ding{53} & \ding{53} & \ding{53} & \ding{51} \\
    \hline
        \makecell[c]{Inferencing\\Time} & \makecell[c]{1.8 ms\\\cite{scholler}} & \makecell[c]{1.8 s\\\cite{social_lstm}} & \makecell[c]{29 ms\\\cite{gatraj}} & \makecell[c]{1.7 ms\\\cite{social_implicit}} & \makecell[c]{12 s\\\cite{flowchain}} & \makecell[c]{10 ms\\\cite{gatraj}} & \makecell[c]{46 ms\\\cite{led}} & \makecell[c]{34 ms\\\cite{flowchain}} & \textbf{0.4 ms} \\
    \hline
    \multicolumn{10}{l}{\ding{53} \footnotesize{no}, \ding{51} \footnotesize{yes}} \\ 
    \multicolumn{10}{l}{$^{1}$\footnotesize{Updated results according to implementation issue \#53 on the Trajectron++ GitHub page}} \\ 
    \end{tabular}
    }
    \vskip -8mm
\end{table}

\section{\large Conclusion}
\label{sec_conclusion}
In compliance with the in~\cref{sec_requirements} described requirements, we present a lightweight method for HTP with a strong focus on application-oriented purposes for autonomous systems. Our method outputs probability distributions, including CL classification for positional uncertainty estimation, and is statistically evaluated to guarantee the correctness of proposed uncertainties. Detailed ablation studies with four datasets demonstrate our approach's performance, flexibility, and usability. The following gives an overview regarding limitations, impacts, and future work. \textbf{Limitations:} Our method focuses on the reliable and fast prediction of human trajectories, including uncertainty assessment with simultaneous use of the minimum available information. Therefore, we solely use past trajectories, excluding context information. Multiple approaches like Trajectron++ demonstrate that interaction handling between road users and using maps can improve the accuracy of predictions. We assume that additional data can increase the accuracy and sharpness of our methods' probabilistic predictions at the cost of increasing resources. Finally, our experiments illustrate that the \textit{Reliability} behavior between the tested datasets can vary, which implies that, in some cases, there is room for further optimizations. \textbf{Impact \& Recommendations:} Our method cannot exceed other HTP methods regarding BoN ADE/FDE accuracy, but it demonstrates good results regarding its relatively simple structure. Our method performs excellently considering statistical validation, i.e., reliability evaluation, uncertainty handling, input horizon flexibility, inferencing time, and hardware requirements. Positional prediction accuracy is only one of multiple essential requirements for autonomous systems. We recommend using probabilistic methods to determine the confidence regions of the predictions and evaluate the reliability to verify the uncertainty. This approach ensures the predictions are statistically sound and can be used safely in downstream tasks like path planning and interaction handling. \textbf{Future Work:} Including additional context information is one of the next logical steps for future development. Maybe the most important one is handling interactions between humans and human-machine behavior. Moreover, sharpness scaling with the forecast horizon should also be investigated.

%
%
\bibliographystyle{splncs04}
\bibliography{egbib}
\end{document}